\documentclass{article} 
\usepackage{iclr2024_conference,times}

\usepackage{amsmath,amsfonts,bm}

\def\eqref#1{equation~\ref{#1}}

\def\1{\bm{1}}

\DeclareMathAlphabet{\mathsfit}{\encodingdefault}{\sfdefault}{m}{sl}
\SetMathAlphabet{\mathsfit}{bold}{\encodingdefault}{\sfdefault}{bx}{n}

\usepackage{hyperref}
\usepackage{url}
\usepackage{graphicx}
\usepackage{pgfplots}
\usepackage{tikzscale}
\usepackage{wrapfig}
\usepackage{subfig}
\usepackage{cleveref}

\usepgfplotslibrary{groupplots}
\pgfplotsset{compat=1.18} 

\title{Bridging Diversity and Uncertainty in Active learning with Self-Supervised Pre-Training}

\author{Paul Doucet, Benjamin Estermann, Till Aczel \& Roger Wattenhofer\\
ETH Zürich\\
\texttt{\{pdoucet,estermann,taczel,wattenhofer\}@ethz.ch} \\
}

\iclrfinalcopy 
\begin{document}

\maketitle

\begin{abstract}

This study addresses the integration of diversity-based and uncertainty-based sampling strategies in active learning, particularly within the context of self-supervised pre-trained models.
We introduce a straightforward heuristic called \textbf{TCM} that mitigates the cold start problem while maintaining strong performance across various data levels.
By initially applying TypiClust for diversity sampling and subsequently transitioning to uncertainty sampling with Margin, our approach effectively combines the strengths of both strategies. 
Our experiments demonstrate that TCM consistently outperforms existing methods across various datasets in both low and high data regimes.

\end{abstract}

\section{Introduction}
Training machine learning models are known to depend on a lot of labeled data.
However, in many settings labeled data is not easy to acquire, but has to be created through expensive manual labeling.
The goal of active learning is to address this challenge by providing a way to select the most informative samples for labeling.
These are the samples for which training a classifier on them increases performance the fastest.

\begin{wrapfigure}{r}{0.41 \textwidth}
    \centering
    \vspace{-10pt}
    \input{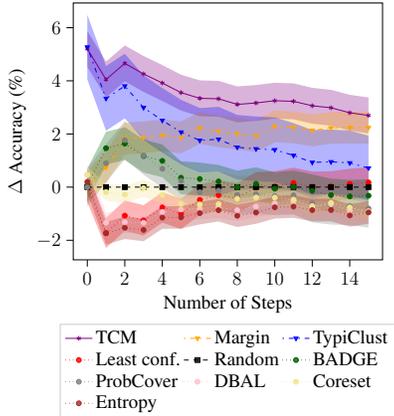}
    \caption{Accuracy improvement compared to random for all baselines and our (TCM) strategy. 
    The accuracy improvement mean and standard deviation is computed over all budget sizes for CIFAR10, CIFAR100 and ISIC2019.
    }
    \vspace{-10pt}
    \label{fig:aggregated_all_datasets}
\end{wrapfigure}

Depending on how many labeled samples the model has already been trained on, the properties of these samples have to change.
In a low-budget setting, diverse and typical samples that cover the complete data distribution are most important.
As the budget increases, the model does not profit from such samples as much anymore.
Instead, once a model has learned the general rules of the data distribution, the most informative samples are now the ones that show where exactly the decision boundaries are.
The issue is that different active learning methods work best in different data budgets and it is not clear when to switch between the methods.

In this work, we provide a new heuristic called TCM on how to combine two such methods, namely TypiClust \citep{Hacohen2022Active} and Margin.
TypiClust shows excellent performance in low data regimes, while Margin excels afterwards.
We specifically analyze the setting where a self-supervised pre-trained backbone model is available.
Such a backbone not only massively increases performance compared to training from scratch, but also the transition dynamics from low to high data regimes simplify.

We show that TCM achieves consistently strong performance, regardless of the labeling budget and the dataset (see \Cref{fig:aggregated_all_datasets}).
It outperforms its underlying methods TypiClust and Margin during the complete training process.
Thanks to the simplicity and effectiveness of our TCM , we provide clear guidelines for practitioners on how to easily use active learning in their own setting.

\section{Related Work}

    In active learning, two primary sampling strategies emerge as critical: diversity-based and uncertainty-based sampling.
    Diversity-based sampling aims to select a representative set of samples that span the entire feature space, thus ensuring broad coverage of the input domain.
    In contrast, uncertainty-based sampling focuses on querying instances for which the model exhibits the highest prediction uncertainty.
    In this way, uncertainty-based sampling aims to refine the model's performance in challenging cases.
    Early on in the training, diversity-based methods tend to perform better as with limited samples it is harder to cover the complete data distribution.
    Further, in that stage, the classifier uncertainty is a weak predictor of hard samples. 
    This is also called the ``cold start problem'' \citep{Mittal2019-oh} of uncertainty-based methods.
    
    \subsection{Self-Supervised Pre-Training}
        Before we introduce any specific active learning methods, we address self-supervised learning.
        In self-supervised learning, a model is trained using a pretext task, allowing it to learn useful representations without explicit external labels. 
        Self-supervised learning complements active learning by learning embeddings before any data is labeled, simplifying both the sample quivering and classifier training.
        Popular and commonly used models include SimCLR \citep{Chen2020Simple} and DINO \citep{Caron2021Emerging}.
        SimCLR uses a contrastive loss, where different views of an image are pushed to have a close representation and views of different images a distant representation.
        The DINO pretext task involves learning data transformation invariant representations, by distilling different views of the same image from a teacher to a student network, where the teacher network is an exponential moving average of the student network.

    \subsection{Uncertainty-Based Active Learning}
        We first briefly introduce relevant uncertainty-based active learning methods.
        Some methods try to quantify model uncertainty based on the output logits of a classifier.
        \textbf{Least confidence} \citep{Lewis1994Sequential} selects samples where the highest class probability is the lowest.
        \textbf{Entropy}  \citep{Joshi2009Multi} measures model prediction uncertainty by the classifier probability distribution and selects samples with the highest entropy.
        \textbf{Margin} selects samples for which the difference between the class probabilities of the most likely two classes is the lowest.
        There also exist methods based on a Bayesian approach.
        \textbf{DBAL} \citep{Gal2017Deep} uses Bayesian convolutional neural networks as the classifier and queries samples based on their highest entropy.
        \textbf{BALD} \citep{Gal2017Deep} also uses Bayesian convolutional neural networks, but in contrast to DBAL, it selects samples that maximize information gained about the model parameters.
        Our TCM strategy builds on top of Margin for uncertainty sampling due to its strong performance and simple design.
    
    \subsection{Diversity-Based Active Learning}
        In the realm of diversity-based active learning, \textbf{Coreset} \citep{Sener2018Active} queries diverse samples by selecting points that form a minimum radius cover of the remaining samples in the unlabeled pool.
        The minimum radius cover ensures that all remaining unlabeled sample has a nearby sample that gets labeled.
        \textbf{ProbCover} \citep{Yehuda2022Active} improves on Coreset by building on top of self-supervised embeddings and selecting samples of high-density regions of the embedding space. 
        While Coreset ensures that it queries samples from the whole distribution, it is likely to select outliers that do not benefit the training.
        In contrast, ProbCover \citep{Yehuda2022Active} samples from uncovered high-density regions, selecting more representative samples.
        \textbf{TypiClust} \citep{Hacohen2022Active} queries diverse samples by first clustering, and then selecting the most typical sample from each cluster. 
        The number of clusters increases by the sampling size at each step, ensuring that there are enough clusters to sample new points from unexplored regions of the embedding space. 
        Typicality is measured by the inverse average distance to other points in the cluster.
        The typical points queried by TypiClust enable strong performance, especially at the beginning of the training process.
        For this reason, our TCM strategy utilizes TypiClust as the initial sampling strategy, thereby avoiding the cold-start problem.

    \subsection{Hybrid Methods}
        Some works in active learning have previously combined diversity and uncertainty-based sampling.
        BADGE \citep{Ash2020Deep} and BatchBALD \citep{Kirsch2019BatchBALD} developed methods to ensure diversity within a batch of uncertain samples.
        SelectAL \citep{Hacohen2023How} on the other hand developed a complex algorithm that automatically detects the current data regime and selects a corresponding diversity or uncertainty-based active learning algorithm.
        However, these methods mostly focused on the case where the classifier is trained from scratch.
        As we show later in this work, when using a pre-trained backbone, there is no need for a complex switching strategy, as the transition point between diversity and uncertainty-based sampling always occurs early on in the training.

\section{Methodology}

    The best querying strategy depends on the already labeled dataset size.
    As can be seen in \Cref{fig:aggregated_all_datasets}, a strong diversity-based method such as TypiClust (Blue) performs strongly in the first few steps of sampling.
    On the other hand, an uncertainty-based method such as Margin (Yellow) shows stronger performance the larger the cumulative budget grows.
    The best and most consistent performance can be achieved when utilizing both methods when they perform the strongest.
    We therefore propose a hybrid sampling strategy combining \textbf{T}ypi\textbf{C}lust and \textbf{M}argin and call this strategy \textbf{TCM}.

    \subsection{Transition Point}
\begin{figure}[!htb]
            \centering
            \includegraphics{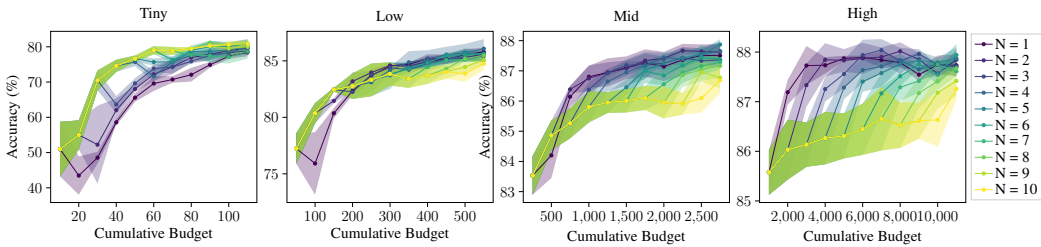}
            \caption{
                Transition point ablation on the CIFAR10 dataset. 
                Switching to Margin in the last step is equal to $N=10$, while only using TypiClust for the initial sampling, and switching to Margin imminently is $N=1$.
            }
            \label{fig:ablation_transition}
        \end{figure}

        To get greater insights into the dynamics of when to transition from TypiClust to Margin, we perform an ablation starting with TypiClust and switching to Margin after $N$ sampling steps.
        The results of this ablation, displayed in \Cref{fig:ablation_transition}, show that the optimal transition point depends on the initial budget and corresponding step size.
        The larger the initial budget, the better it is to quickly switch from TypiClust to Margin to achieve consistently strong performance.
        
    \subsection{Step Size}
\begin{figure}[!htb]
            \centering
            \includegraphics{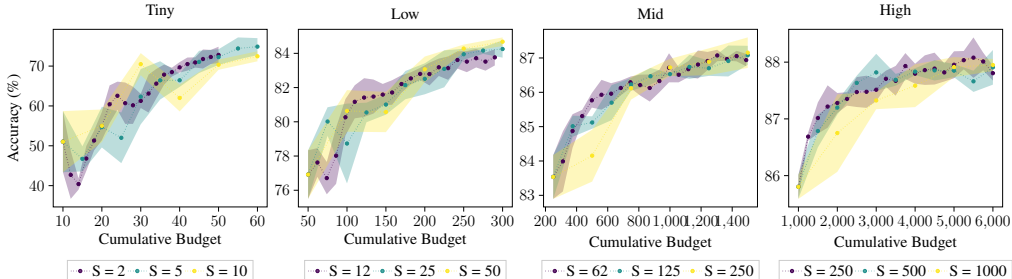}
            \caption{
                Step size ablation for TCM on the CIFAR10 dataset. 
                For each regime, we evaluate three different step sizes $S$.
                Overall, there is no clear performance difference between the step sizes.
            }
            \label{fig:ablation_steps}
        \end{figure}

        We further analyze the effect of the step size for the Margin part of TCM.
        Thanks to the design of TypiClust, its performance is mostly independent of step size.
        For this reason, we only perform detailed analysis on Margin, after having selected the initial budget with TypiClust.
        By just selecting a batch of samples depending on the class probabilities, Margin might select a lot of similar samples if the step size is too big.
        At the same time, too small of a step size is not practical, as the model would need to be retrained too often.
        For this reason, we perform an ablation on the effect of the step size on the performance of TCM, shown in \Cref{fig:ablation_steps}.
        Surprisingly, the results show that overall, there is no clear difference in performance for different step sizes.
        A large step size for the Mid and High budget might have a slightly negative impact, however towards the end of the cumulative budget, the difference disappears.
        The implications of these results are strongly positive, as they indicate that the step size of TCM can be adapted to other needs such as the availability of experts for labeling data and computational resources.

        Based on our ablations, we devise the following simple heuristic for TCM.
        We use a step size equal to the size of the initial budget of each setting. 
        In the Tiny and Low settings, we perform 3 steps of TypiClust before switching to Margin.
        In the Medium setting, we perform 2 steps of Typiclust and in the High setting, we perform a single step.
        
        However, based on our ablations we can also provide a broader rule of thumb for an active learning practitioner.
        In a new setting, we suggest using a total budget of roughly 20 times the number of categories for TypiClust and then switching to Margin, with no strict constraint on step size.

\section{Experimental Evaluation}
    
    \subsection{Setup}
        Today's active learning research has an inconsistent landscape with many contradicting results. 
        For this reason, we follow the evaluation framework proposed by \citet{Luth2023Toward} to have a fair and rigorous comparison to all active learning baselines.
        In particular, we tune the hyperparameters on the CIFAR10 \citep{krizhevsky2009learning} dataset and use all other datasets as rollout datasets. 
        We evaluate the datasets CIFAR10 and CIFAR100 \citep{krizhevsky2009learning} and ISIC2019 \citep{Codella2017ISIC}, but also consider the long-tail versions of the two datasets, denoted with LT \citep{Cao2019Learning}.
        These long-tail versions feature class imbalance, where the number after LT indicates the ratio between the most common and the least common class, combined with an exponential decay in sample sizes for the other classes.
        We perform all our experiments on 4 different data budget sizes. 
        These include small, medium, and large as defined by \citeauthor{Luth2023Toward}, as well as the tiny budget size with the initial sample size equal to the number of classes.
        Further, we consistently use a step size equal to the number of initial samples for all budgets.
        Some sampling methods such as ProbCover, Typiclust and our proposed TCM rely on self-supervised pre-trained representations.
        For all these methods, we use the same representations.
        Specifically, for CIFAR10, CIFAR10-LT5, CIFAR10-LT10, CIFAR100, CIFAR100-LT5, CIFAR100-LT10 we use the SimCLR features provided by \citeauthor{Hacohen2022Active}.
        For ISIC2019, we train a DINO model from scratch and use the features of the model.

\begin{figure}[h]
        \centering
        \includegraphics{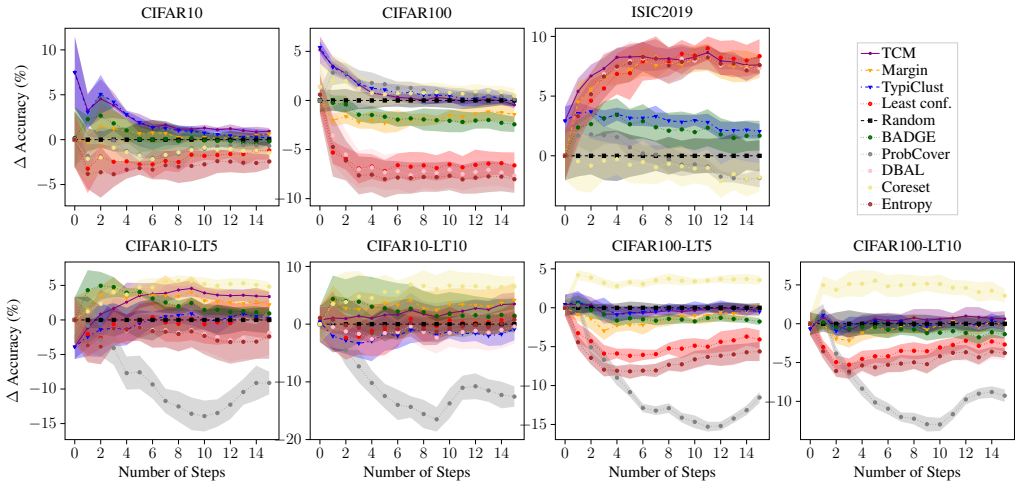}
        \caption{Accuracy improvement compared to random for all baselines and our (TCM) strategy. 
        The accuracy improvement mean is computed over all 4 budget sizes tiny, small, medium, and large.
        Standard deviation is aggregated with respect to the random seed.
        The top row shows the main evaluated datasets, while the bottom row shows an ablation on the imbalanced versions of CIFAR10 and CIFAR100.
        For all imbalanced datasets, reported accuracy is balanced by computing the average of recall obtained for each class.
        TCM shows consistently strong performance for all datasets, even for datasets for which TypiClust or Margin on their own show suboptimal performance.
        Coreset shows strong performance on the LT datasets.
        Unfortunately, this performance does not transfer to the real-life imbalanced dataset ISIC2019.}
        \label{fig:results}
    \end{figure}

        To underpin the efficacy of our proposed active learning framework, we leverage the same pre-trained representations for training our classifier.
        For all experiments conducted within this work, we utilize the selected backbone as a foundational feature extractor.
        On top, we train a linear prediction head, fine-tuning the model to the labels gained during the active learning process.
        This approach allows us to harness the rich representational capacity of the pre-trained models, leading to a much stronger performance of the classifier especially in a low-data regime.
        Furthermore, by using publicly available pre-trained model embeddings, the classifier training following each sampling step demands significantly fewer computational resources.
        Due to the above mentioned advantages of using a pre-trained backbone and considering that TypiClust relies on such a backbone, we do not perform any experiments where we train a classifier from scratch.
        We fix all training hyperparameters for classifier training for all datasets, budgets, and querying strategies.
        For all experiments, we plot the mean and standard deviation of runs with 3 seeds.
        While we have already presented an aggregation over all datasets and budget sizes in \Cref{fig:aggregated_all_datasets}, we present specific results for each evaluated dataset in \Cref{fig:results}.
    
    \subsection{Results}
        Our results showcase the consistent and strong performance of TCM compared to other baselines.
        This performance gain transfers to the imbalanced long-tail versions of CIFAR10 and CIFAR100.
        It can be seen that TCM performs well even if either TypiClust or Margin do not perform well on their own.
        In CIFAR10-LT5 as well as in CIFAR100-LT10, TCM performs better than both its underlying methods, showcasing the potential of the combination.
        On the main datasets CIFAR10, CIFAR100 and ISIC2019, Coreset shows performance that is worse than selecting samples at random, but shows strong performance in the LT ablation.
        We hypothesize that this could be because, for the LT datasets, the backbone was pre-trained on the balanced version of the dataset.
        This is not the case for ISIC2019, where the DINO backbone was trained entirely on ISIC2019.
        Other methods such as ProbCover, DBAL, or Least confidence seem to struggle a lot with providing consistent performance in all data regimes and datasets.
        Furthermore, our results show that when using a pre-trained backbone, more complex methods such as SelectAL do not seem necessary, because the transition point occurs much earlier compared to training from scratch.
        This fact is exploited by TCM and allows for its strong and consistent performance for all budget sizes.
        Switching between diversity and uncertainty-based strategies after the few first iterations consistently outperforms other methods.

\section{Conclusion}
    In this work, we have shown that when training a classifier using a pre-trained backbone, the transition point from diversity to uncertainty-based active learning methods occurs early.
    Based on these results, we present TCM, a simple, yet effective hybrid strategy that outperforms all other methods when compared on a wide range of data regimes and datasets.
    By using TypiClust for the first few steps and then switching to Margin, TCM selects informative instances in both low and high data regimes.
    Using the simple heuristics laid out by TCM, practitioners can apply active learning easily and effectively to their use case.
    
\newpage

\bibliography{iclr2024_conference}
\bibliographystyle{iclr2024_conference}

\end{document}